# Enhancing IoT Cyber Attack Detection in the Presence of Highly Imbalanced Data

Md. Ehsanul Haque, Md. Saymon Hosen Polash, Md Al-Imran
Sanjida Simla, Md Alomgir Hossain, Sarwar Jahan
Department of Computer Science and Engineering
East West University, Dhaka, Bangladesh
Emails: {ehsanulhaquesohan758@gmail.com, polash3063013@gmail.com, al.imran@ewubd.edu,
sanjida.simla55@gmail.com, malomgirh01@gmail.com, sjahan@ewubd.edu}

*Abstract*—Due to the rapid growth in the number of Internet of Things (IoT) networks, the cyber risk has increased exponentially, and therefore, we have to develop effective IDS that can work well with highly imbalanced datasets. High rate of missed threats can be the result as traditional machine learning models tend to struggle in identifying attacks as normal data volume is so much higher than the volume of attacks. For example, the data set used in this study reveals a strong class imbalance with 94,659 instances of the majority class and only 28 instances of the minority class, in which determining rare attacks accurately is quite challenging. The challenges presented in this research are addressed by hybrid sampling techniques designed to drive data imbalance detection accuracy in IoT domains. After doing so, we then evaluate the performance of several machine learning models such as Random Forest, Soft Voting, Support Vector Classifier (SVC), K-Nearest Neighbors (KNN), Multi-Layer Perceptron (MLP) and Logistic Regression with respect to the classification of cyber-attacks accurately. The obtained results indicate that the Random Forest model achieved the best performance value of 0.9903 of Kappa score, 0.9961 of test accuracy and 0.9994 of AUC. It also shows strong performance in the Soft Voting model, with an accuracy of 0.9952 and AUC of 0.9997, while the latter is an indication of the benefits of combining models' prediction. Overall, this work has shown the great benefit of hybrid sampling combined with robust model selection and feature selection to deliver a dramatic increase in of IoT security against cyber-attack, an important factor for implementing security in environments with strongly imbalanced data.

## I. INTRODUCTION

The Internet of Things (IoT) devices have become rapidly proliferating devices to transform enough of industries and make the smooth communication between billions of connected devices. By 2030, almost twice as many IoT devices (32.1 billion) will be used worldwide when compared to the number of 15.9 billion in 2023 [1]. As connectivity explodes data is being generated at truly massive scale. According to International Data corporation, by 2025, total data generated across the world will be around 175 zettabytes (ZB) in which approximately 80 ZB is contributed by IoT devices [2].

However, as adoption of IoT has surged, so has the security challenge. Millions of cyber attacks targeting IoT devices are being reported, with an increase from 639 Million in 2020 to 1.51 Billion in the first half of 2021 [3]. IoT networks present a unique set of characteristics that require traditional cybersecurity measures to adapt to, often through highly imbalanced data. While existing detection systems suffer tremendously from the overwhelming benign traffic, attacks go undetected.

The research examines a framework that aims to achieve high precision and recall in IoT cyber-attack detection for imbalanced datasets having 94,659 majority class instances and 28 minority class instances. Put forward to address the imbalanced data problems like low precision and recall, fine-tuned machine learning model is used in this experiment along with hybrid sampling strategies to boost detection precision while suppressing false detections and false negatives [5] [11].

In terms of the contribution, this study specifically contributes to analyzing the data coming out of IoT devices and having a major impact on IoT security and resulting safety.

The remaining sections of the study are divided as follows: Section II discusses the Related Works, Section III features the Methodology, Section IV introduces the Evaluation Metrics, Section V presents the Results and Discussion, Section VI introduces the Proposed Approach and Section VII concludes.

## II. RELATED WORK

SVM, GBDT and RF were used by Muhammad et al. [6] for the purpose of detecting IoT attack into the NSL-KDD dataset. Next to other models, RF achieved the highest accuracy of 85.34%. Nevertheless, the accuracy is not so high for the IoT security demands. In general, found gaps were model optimization, exploration of the new technique domain, and testing on the newer IoT datasets.

HMM and SVM were proposed as real time anomaly detection framework by Dr. Vaishali et al [7] for Healthcare IoT. The model has achieved 98.66% accuracy in PhysioNet Challenge 2017 dataset and outperformed Naïve Bayes and LSTM. Nevertheless, its versatility in various IoT devices and real-world mixed data is unknown. More research is also needed to make this scalable and user acceptable in clinical applications.

FusionNet, introduced by Dheyaaldin [8], is a new ensemble model that consists of ensemble of RF, KNN, SVM, and MLP for anomaly detection. On two datasets, it produced accuracy of 98.5% and 99.5% respectively indicating strong classification. Nevertheless, the study was done on small sample sizes with a multi-class classification technique. In the future work it will be worth improving the anomaly detection in real world applications on larger datasets containing more classes.

In the study done by Maryam et al [9], ML based IoT anomaly detection was analyzed with the CIC IoT dataset. Both binary and multi-class classifications on their RF model reached 99.55% accuracy. Nevertheless, difficulties in correlation and the reduction of dimensionality of the data necessary for the achievement of an enhanced performance were pointed out. On top of that, the study also found that RF has a difficult time to differentiate some attack types of Recon and Spoofing.

ML intrusion detection for IoT security was explored in the paper Supongmen Walling et al. [10]. Thus, feature selection using ANOVA and Pearson correlation has been employed to improve anomaly based NIDS performance. Feature selection was shown to be critical by showing 99.6% accuracy and 97.7% for SVM and RF respectively in experiments. The nature of threat to modern cyberspace calls for adaptive IDS models in response to fast evolving cyber threats.

Within the SDN-WISE IoT framework, Jalal et al. [12] proposed the use of a DT based detection module in enhancing network security. On simulated DDoS traffic, their model reached 98.1% accuracy with only 30% of memory and CPU resources. It shows an efficient way of securing IoT regarding computation. We leave this work for future research, and the scalability and effectiveness of it under diverse IoT attack vectors should be evaluated.

Bot-IoT dataset was used by Jadel et al. [13] to analyze the vulnerabilities of the IoT network. They compared the ML models to detect Probing, DoS attacks and KNN has highest accuracy at 99%. However, real time applicability was limited because of high training time. The paper illustrates this tradeoff between detection accuracy and efficiency in IoT security.

ML based anomaly detection in IoT networks was investigated by Nicolas Alin [14]. The accuracy of RF, NB, MLP, SVM and A, is 99.5%, the best result achieved is with RF. Although they showed precision, recall, and F1 scores, there was improvement needed in attack classification. Yet, there is still some improvement that could be done for model robustness and for solving the IoT specific security challenges.

## III. METHODOLOGY

The following methods are used for this experiment to detect IOT cyber attacks in ROT-IOT22 dataset in a efficient way.

### A. Data Collection

The dataset used for RT-IOT2022 comes from the UCI Machine Learning Repository [4] which consists of 123,117 instances and 83 features (81 numerical and 2 categorical). This contains network traffic data with 12 IoT cyber attack classes, for the case study of cybersecurity threats, as well as in improving machine learning algorithms to attack identification. Distribution of 12 attack types is given in Table I.

TABLE I: Distribution of Cyber Attacks in the Dataset

| Attack | Count | Attack | Count |
| --- | --- | --- | --- |
| SYN Hping | 94659 | UDP Scan | 2590 |
| Thing Speak | 8108 | Tree Scan | 2010 |
| ARP Poisoning | 7750 | NMAP OS | 2000 |
| MQTT Publish | 4146 | TCP Scan | 1002 |
| DDOS Slowloris | 534 | Wipro Bulb | 253 |
| Brute Force | 37 | FIN Scan | 28 |

### B. Preprocessing

Preprocessing of data is much important for giving good quality input in machine learning tasks for getting better and reliable result. This also helps clean and structure raw data, which significantly improves performance of our model. The main steps in data preprocessing include:

*1) Encoding Categorical Columns:* In the dataset, two categorical columns terms were converted into numerical value using label encoding. This helps in simplifying the process of models analyzing data.

*2) Outlier Removal:* The Z-score method identifies and removes outliers by identifying the points of data that are very different from the mean. It helps to keep the dataset clean and reliable, also maintaining its quality for further analysis.

*3) Standardization:* Standardized is performed to ensure all of the features to have mean zero and variance one. This allows for the process of improving model performance by making sure each feature is both contributed to equally, and best detects.

*4) Feature Selection:* There are 83 columns in the dataset, of which includes features and the target variable. Thus, Recursive Feature Elimination (RFE) was administered in order to improve the performance of the model as well as lower the computational cost. RFE eliminates features one by one through this step, leaving only the most significant ones. After the RFE process, 20 features were selected which were most important in the model, thus having the highest efficiency of the model without the risk of overfitting.

*5) Training and Split:* A split was made in the dataset to train and test with an 80:20 ratio. By doing so model is learning effectively from the training data, but from unbiased the test on unseen data.

*6) Data Balancing:* The class imbalance issue was resolved by Hybrid sampling [5]. For the majority class (class 2) it reduced to 10,000 instances, and for minority classes (classes 0, 1, 3–10) it is increased to 10000 instances each also. Such a balanced distribution thus allows the model to learn effectively from majority as well as minority classes.

Figure 2 shows the class distribution before and after applying hybrid sampling.

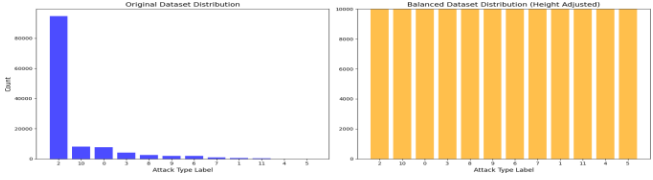

Fig. 1: Before and After Balancing

*7) Model Selection:* In this study, Random Forest, Soft Voting, SVC, KNN, MLP, and Logistic Regression are trained on balanced dataset. This approach enabled each model to tackle the imbalanced IoT data fairly and allows it to fairly assess the ability of models to detect cyber attacks.
This makes effective IoT cyber attack detection possible even in highly imbalanced datasets by following these steps. The proposed approach improves the accuracy and reliability of identifying different attacking styles in IoT networks.

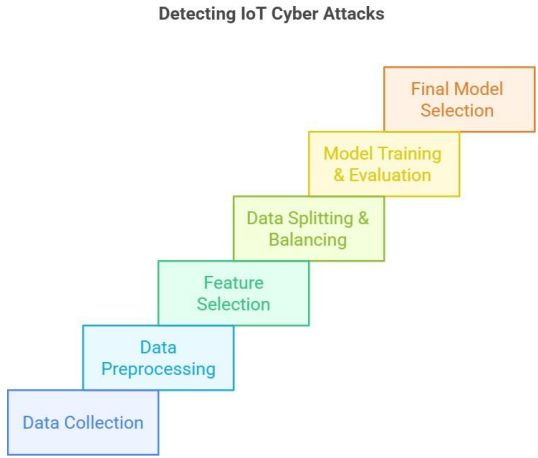

Fig. 2: Workflow Diagram

## IV. RESULTS AND DISCUSSIONS

The results on the balanced dataset obtained from the various machine learning models for IoT cyber-attack detection are presented in this section. The model settings were fine-tuned using Grid Search so as to improve performance on training and testing. All models assessed comprehensively and described their performance on several metrics.

To evaluate the model effectiveness, the accuracy, precision, recall, F1-score, Kappa Score, AUC-ROC were taken into consideration for a more comprehensive assessment. Finally, the confusion matrix was interpreted to further analyze errors of classification. Comprising these metrics applied on the proposed models, these provided the basis for the evaluation of their robustness and were used to rigorously analyze the capacity of the models to detect cyber threats in IoT environments.

### A. Training Performance

The training performance metrics and cross-validation scores from various machine learning models used to detect IoT cyber attacks are presented in Table II. The table displays the accuracy results along with the training execution time and cross-validation score (CV score) of each model. Random Forest delivers the best detection accuracy rate at 0.9956 followed by a training duration of 18.60 seconds. The Support Vector Classifier (SVC) produces 0.9741 accuracy yet needs 244.20 seconds for its training process. The K-Nearest Neighbors (KNN) model exhibits the best accuracy levels of 0.9969 and the fastest training duration of 0.09 seconds. The training process for Multilayer Perceptron (MLP) takes 165.24 seconds to reach an accuracy level of 0.9941. The model performance of Logistic Regression (LR) reaches a training time of 55.70 seconds and shows an accuracy of 0.9754. The Soft Voting algorithm achieves excellent accuracy of 0.9939 but it demands the longest training time at 501.36 seconds.

For real-time applications, the long computational processing time of 244.20 seconds for SVC and 501.36 seconds for Soft Voting limits their efficiency despite having accuracies of respectively 0.9741 and 0.9939. Also, it is observed from the table that all trained models demonstrated their capability to generalize well through their close cross-validation accuracy match with training accuracy. The models demonstrate reliable behavior on new data sets throughout their validation process because of their consistent performance.

The below figures are added to provide a better understanding of the model performance. Figure 3 presents the learning curves for the Random Forest (RF) and Support Vector Classifier (SVC) models. Figure 4 illustrates the learning curves for the K-Nearest Neighbors (KNN) and Multi-Layer Perceptron (MLP) models. Lastly, Figure 5 showcases the learning curves for Logistic Regression (LR) and the Soft Voting ensemble method.

TABLE II: Model Performance Metrics and Cross-Validation Scores

| Model | Accuracy | Training Time (s) | CV Score |
|---|---|---|---|
| RF | 0.9956 | 18.60 | 0.9938 |
| SVC | 0.9741 | 244.20 | 0.9723 |
| KNN | 0.9969 | 0.09 | 0.9953 |
| MLP | 0.9941 | 165.24 | 0.9932 |
| LR | 0.9754 | 55.70 | 0.9744 |
| Voting (soft) | 0.9939 | 501.36 | 0.9927 |

The learning curves for all models provided in Figures 3, 4, and 5 respectively depict high predictive accuracy learning from the training data. Among all, Random Forest (RF) and K-Nearest Neighbors (KNN) show the least overfitting and hence, great accuracy. SVC and MLP also allow high accuracy, while several others, such as Logistic Regression and Soft Voting, show how the best predictions are arrived at when models are combined. In total, the findings reveal that all models indeed retain essential characteristics of the data.

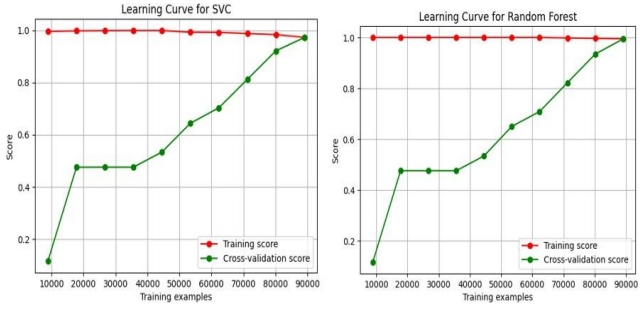

Fig. 3: Learning curves of Random Forest (RF), Support Vector Classifier (SVC)

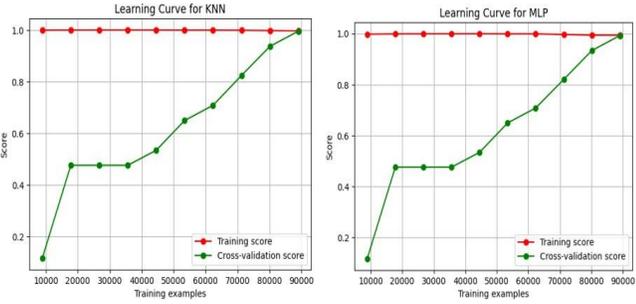

Fig. 4: Learning curves of K-Nearest Neighbors (KNN), Multi-Layer Perceptron (MLP)

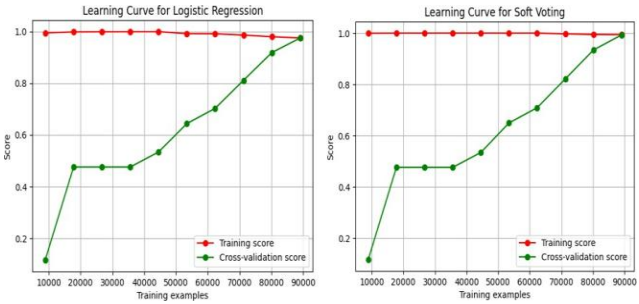

Fig. 5: Learning curves of Logistic Regression (LR), and Soft Voting

### B. Testing Performance

Table III demonstrates the comparison of testing performance of different Machine Learning algorithms applied for the cyber attack identification IoT. The RF model detects intrusions as good or better than any model as its test accuracy rate is 0.9961 and test AUC is 0.9994. The benefit of the Soft Voting model is also notable at which test accuracy is 0.9952 and AUC is 0.9997, evidencing the good cooperation of classifiers. The Support Vector Classifier (SVC) model can be used with a test accuracy of 0.9811 and AUC of 0.9980 and shows its ability. Similarly, KNN achieves very good test accuracy of 0.9952 and and AUC of 0.9737 which ensures its reliability on this domain. The test accuracy of such the Multi Layer Perceptron (MLP) model is as high as 99.57% and its AUC is 99.64% Finally Logistic Regression (LR) achieved a test accuracy of 0.9828 and an AUC value of 0.9963. Finally, the performance of the presented models in network intrusion detection are given and the results are utilized for further applications.

TABLE III: Model Performance Metrics

|  | RF | SVC | KNN | MLP | LR | Voting |
|---|---|---|---|---|---|---|
| **Accuracy** | 0.9961 | 0.9811 | 0.9952 | 0.9957 | 0.9828 | 0.9952 |
| **AUC** | 0.9994 | 0.9980 | 0.9737 | 0.9964 | 0.9963 | 0.9997 |

To further validate the results Precision, recall, and F1 measures were used to evaluate the classification performance of each model, and these metrics pointed to their usefulness in differentiate classes displayed on table IV .The table IV summarizes the performance of various machine learning models using key evaluation metrics: The following corresponding values of Macro Precision (MP), Weighted Precision (WP) , Macro Recall (MR), Weighted Recall (WR), Macro F1 Score (MF1) and Weighted F1 Score (WF1) are calculated. Random Forest delivers excellent performance across all metrics based on its near-perfect weighted metrics which include WP = 1.00, WR = 1.00, and WF1 = 1.00. Support Vector Classifier (SVC) presents a mix of performance factors with Macro Precision at 0.88 and Macro Recall at 0.92 and high Weighted Precision at 0.98 and Weighted F1 Score at 0.98. K-Nearest Neighbors (KNN), Multi-Layer Perceptron (MLP), and Voting display matched balanced outcomes using WP, WR and WF1 metrics reaching 1.00 value. The results from Logistic Regression indicate acceptable performance based on MP = 0.79 and MR = 0.92. RF together with MLP and Voting displayed the best performance when dealing with imbalanced datasets since their precision and recall outcomes were both high.

TABLE IV: Performance Metrics for Different Models

| Model | MP | WP | MR | WR | MF1 | WF1 |
|---|---|---|---|---|---|---|
| RF | 0.95 | 1.00 | 0.93 | 1.00 | 0.94 | 1.00 |
| SVC | 0.88 | 0.98 | 0.92 | 0.98 | 0.82 | 0.98 |
| KNN | 0.87 | 1.00 | 0.94 | 1.00 | 0.89 | 1.00 |
| MLP | 0.90 | 1.00 | 0.94 | 1.00 | 0.91 | 1.00 |
| LR | 0.79 | 0.99 | 0.92 | 0.98 | 0.82 | 0.98 |
| Voting | 0.87 | 1.00 | 0.94 | 1.00 | 0.90 | 1.00 |

The performance of each model in terms of misclassified and correctly classified data is presented in the confusion matrices from Figure 6 to 8. The performance of the Random Forest (RF) model is better than the Multi-Layer Perceptron (MLP) model, and as seen in Figure 6, there are fewer misclassifications. The Support Vector Classifier (SVC) slightly beats Logistic Regression (LR) in performance, having fewer false positives and false negatives, as seen in Figure 7. Finally, in Figure 8, it presents the relative competitiveness of the Voting Classifier and the K-Nearest Neighbors (KNN), and the Voting Classifier has a slightly lower misclassification rate. Overall, all three figures show that Random Forest (RF) indeed has the most consistent and reliable classification results in IoT cyber attack detection.

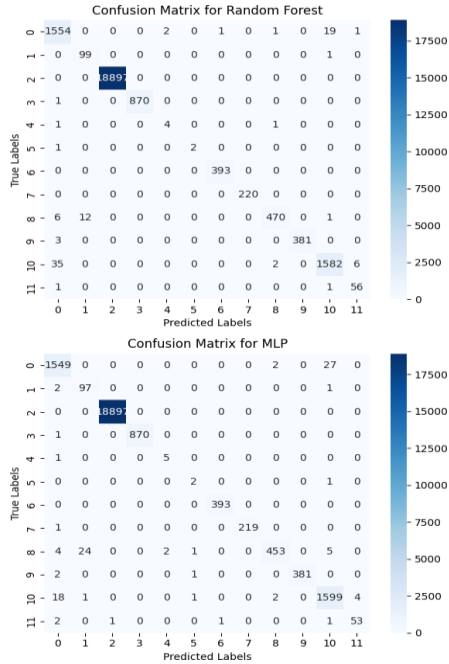

Fig. 6: Confusion Matrix of RF & SVC

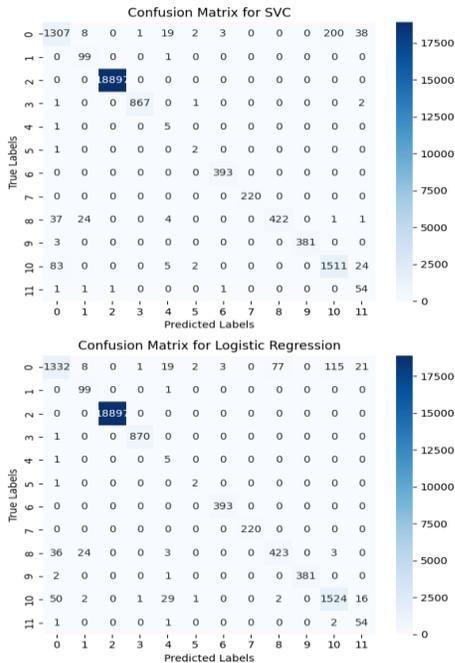

Fig. 7: Confusion Matrix of SVC & LR

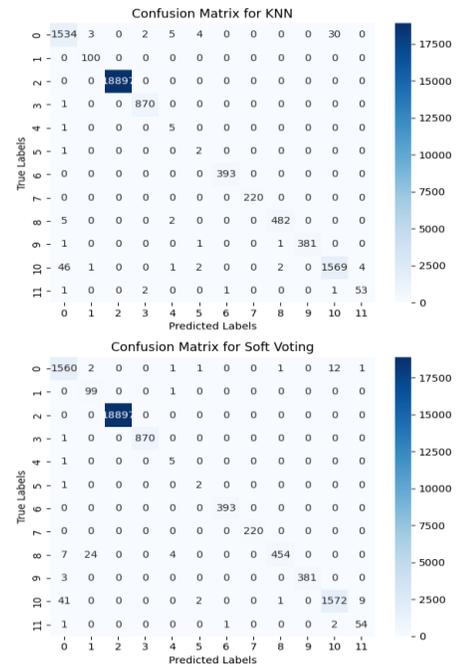

Fig. 8: Confusion Matrix of KNN & Voting

*C. Kappa Assessment*

The previous section discusses the evaluation metrics AUC, accuracy, precision, recall, and the confusion matrix. The Kappa test was conducted to validate the results and is displayed in Table V. The Kappa scores shown in Table V and it is found that Random Forest model obtains highest Kappa score of 0.9903, surpassing other models with the ability to correctly classify instances. The KNN and MLP models also did very well (0.9880) and (0.9892). Scores of a moderate 0.9527 for SVC and 0.9570 for Logistic Regression respectively indicates lower performance for these models. Soft Voting classifier achieved a Kappa score of 0.9881 suggesting that this classifier combined predictions of multiple models for a better accuracy. However, these results show overall that Random Forest and Soft Voting ensemble methods perform well in this classification task.

TABLE V: Kappa Scores for Different Models

| Model | RF | SVC | KNN | MLP | LR | Voting |
|---|---|---|---|---|---|---|
| Kappa | 0.9903 | 0.9527 | 0.9880 | 0.9892 | 0.9570 | 0.9881 |

*D. Proposed Approach*

Based on the above analysis, it is see that the Random Forest model is outperforming other models according to accuracy, AUC, Kappa and finally minimizing false positive and false negative. Moreover, it is find that the Random Forest model is more efficient, with reduced less training time compared to the other applied models which is only 18.6 seconds. Therefore, Random Forest is suggested as the proposed model for cyber attacks detection from the RT IoT 22 dataset.

Below are the hyperparameters used for the **Random Forest** model to ensure its optimal performance given in table VII:

TABLE VI: Random Forest Hyperparameters

| Hyperparameter | Value |
|---|---|
| random_state | 42 |
| n_estimators | 100 |
| max_depth | 10 |
| min_samples_split | 5 |
| min_samples_leaf | 2 |

## V. Comparative Analayis with previous study

TABLE VII: Summary of Study Datasets and Best Performing Models

| Study | Best Model | Accuracy |
|---|---|---|
| [7] | SVM | 98.66% |
| [8] | FusionNet | 99.52% |
| [9] | RF | 99.55% |
| [10] | SVM | 99.60% |
| This Study | RF | 99.61% |

The proposed approach outperformed the prior IOT cyber attack detection study and achieved an accuracy of 99.61% with help of Random Forest. This improvement proves the effectiveness of using an efficient sampling strategy according to the optimized models thereby presenting key findings in increasing IoT security against new emergent cyber threats.

## VI. Conclusion

The research examines how to identify cyber-attacks present in unbalanced datasets of IoT networks. Hybrid sampling together with machine learning model testing enabled the Random Forest model to reach an accuracy of 99.61% and a Kappa score of 0.9903 indicating the value of both model selection and data balancing.

The upcoming research direction includes deep learning attack detection methodologies and web-based detection API creation together with IoT network-based evaluation and additional feature selection optimization. The model's robustness alongside generalization abilities could be increased by implementing adaptive learning systems that apply to various IoT datasets for testing purposes.

## References

[1] Lionel Sujay Vailshery, *Number of IoT connections worldwide 2022-2033*, Statista, September 11, 2024. Available: https://www.statista.com/statistics/1183457/iot-connected-devices-worldwide/.

[2] Unknown, *The IoT sweet spot – how much data is too much data?*, ERP Today. Available: https://erp.today/the-iot-sweet-spot-how-much-data-is-too-much-data/. Accessed: October 28, 2024.

[3] Cyrus, Callum, *IoT Cyberattacks Escalate in 2021, According to Kaspersky*, IoT World Today. Available: https://www.iotworldtoday.com/security/iot-cyberattacks-escalate-in-2021-according-to-kaspersky. Accessed: October 28, 2024.

[4] S., B. & Nagapadma, R. (2023). RT-IoT2022 [Dataset]. UCI Machine Learning Repository. https://doi.org/10.24432/C5P338.

[5] Chris Seiffert, Taghi M. Khoshgoftaar, and Jason Van Hulse, "Hybrid sampling for imbalanced data," in *2008 IEEE International Conference on Information Reuse and Integration*, pp. 202-207, 2008. DOI: 10.1109/IRI.2008.4583030.

[6] Hyder, Muhammad Faraz & Nazir, Waseemullah & Farooq, Muhammad & Anwer, Maryam & Khan, Shariq. (2021). Attack Detection in IoT using Machine Learning. Engineering, Technology & Applied Science Research. 11. 7273-7278. 10.48084/etasr.4202.

[7] Vaishali V. Raje, Shalini Goel, Sujata V. Patil, Mahadeo D. Kokate, Dhiraj A. Mane, Santosh Lavate, *Realtime Anomaly Detection in Healthcare IoT: A Machine Learning-Driven Security Framework*, J. Electrical Systems, **19**(3), 192-202, 2023. DOI: https://doi.org/10.52783/jes.700

[8] Dheyaaldin, A. (2024). A comparative study of anomaly detection techniques for IoT security using adaptive machine learning for IoT threats. *IEEE Access*, 12, 14719-14730. DOI:10.1109/ACCESS.2024.3359033

[9] Khan, M.M., Alkhathami, M. Anomaly detection in IoT-based healthcare: machine learning for enhanced security. Sci Rep 14, 5872 (2024). https://doi.org/10.1038/s41598-024-56126-x

[10] Walling, Supongmen and Lodh, Sibesh. "Network intrusion detection system for IoT security using machine learning and statistical based hybrid feature selection." *SECURITY AND PRIVACY*, 2024;e429. doi: https://doi.org/10.1002/spy2.429.

[11] M. E. Haque, A. Hossain, M. S. Alam, A. H. Siam, S. M. F. Rabbi, and M. M. Rahman, "Optimizing DDoS Detection in SDNs Through Machine Learning Models," in *Proc. 2024 IEEE 16th Int. Conf. on Computational Intelligence and Communication Networks (CICN)*, Bhopal, India, 2024, pp. 426–431, doi: 10.1109/CICN63059.2024.10847458.

[12] J. Bhayo, S. A. Shah, S. Hameed, A. Ahmed, J. Nasir, and D. Draheim, "Towards a machine learning-based framework for DDOS attack detection in software-defined IoT (SD-IoT) networks," *Engineering Applications of Artificial Intelligence*, vol. 123, p. 106432, 2023. doi: 10.1016/j.engappai.2023.106432.

[13] J. Alsamiri and K. Alsubhi, "Internet of Things Cyber Attacks Detection using Machine Learning," *International Journal of Advanced Computer Science and Applications (IJACSA)*, vol. 10, no. 12, 2019. [Online]. Available: https://thesai.org/Downloads/Volume10No12/Paper_80-Internet_of_Things_Cyber_Attacks_Detection.pdf

[14] N.A. Stoian, *Machine Learning for anomaly detection in IoT networks: Malware analysis on the IoT-23 data set*, July 2020. Available at: http://essay.utwente.nl/81979/